\documentclass[conference]{IEEEtran}
\IEEEoverridecommandlockouts
\usepackage{cite}
\usepackage{amsmath,amssymb,amsfonts}
\usepackage{algorithmic}
\usepackage{graphicx}
\usepackage{textcomp}
\usepackage{xcolor}
\usepackage{float}
\usepackage{algorithm}
\usepackage{amssymb}
\usepackage{booktabs} 
\def\BibTeX{{\rm B\kern-.05em{\sc i\kern-.025em b}\kern-.08em
    T\kern-.1667em\lower.7ex\hbox{E}\kern-.125emX}}
\begin{document}

\title{Partially Shared Semi-supervised Deep Matrix Factorization with Multi-view Data\\
}

\author{\IEEEauthorblockN{Haonan Huang\IEEEauthorrefmark{2}, Naiyao Liang\IEEEauthorrefmark{2}\IEEEauthorrefmark{3}, Wei Yan\IEEEauthorrefmark{2}, Zuyuan Yang\IEEEauthorrefmark{2}\IEEEauthorrefmark{5}, Weijun Sun\IEEEauthorrefmark{2}\IEEEauthorrefmark{4}}
\IEEEauthorblockA{\IEEEauthorrefmark{2}\textit{Guangdong Key Laboratory of IoT Information Technology, Guangdong University of Technology,
Guangzhou, China}}
\IEEEauthorblockA{\IEEEauthorrefmark{3}\textit{Key Laboratory of iDetection and Manufacturing-IoT, Ministry of Education, Guangzhou, China}}
\IEEEauthorblockA{\IEEEauthorrefmark{4}\textit{Guangdong-Hong Kong-Macao Joint Laboratory for Smart Discrete Manufacturing, Guangzhou, China}}
\IEEEauthorblockA{\IEEEauthorrefmark{5}\textit{Author to whom any correspondence should be addressed.}}
\{mrhaonan, naiyaogdut\}@aliyun.com, helloyanwei@163.com, \{yangzuyuan, gdutswj\}@gdut.edu.cn
}
\maketitle

\begin{abstract}
Since many real-world data can be described from multiple views, multi-view learning has attracted considerable attention. Various methods have been proposed and successfully applied to multi-view learning, typically based on matrix factorization models. Recently, it is extended to the deep structure to exploit the hierarchical information of multi-view data, but the view-specific features and the label information are seldom considered. To address these concerns, we present a partially shared semi-supervised deep matrix factorization model (PSDMF). By integrating the partially shared deep decomposition structure, graph regularization and the semi-supervised regression model, PSDMF can learn a compact and discriminative representation through eliminating the effects of uncorrelated information. In addition, we develop an efficient iterative updating algorithm for PSDMF. Extensive experiments on five benchmark datasets demonstrate that PSDMF can achieve better performance than the state-of-the-art multi-view learning approaches. The MATLAB source code is available at https://github.com/libertyhhn/PartiallySharedDMF.
\end{abstract}


\section{Introduction}
In practical applications, real-world data can be described from different views, that is the so called multi-view data. For instance, an image can be described by several characteristics, e.g., shape, color, texture and so on. Because multi-view representation learning can exploit the implicit-dependent structure of multiple views and improve the performance of the learning tasks. In recent years, multi-view representation learning has attracted increasing research attention in machine learning \cite{MVML,survey1,survey2}.

Over the past decade, lots of researches on multi-view learning have emerged. In particular, Non-negative Matrix Factorization (NMF) \cite{NMF,AnsNMF} as one of the most popular high-dimensional data processing algorithms, have been widely used for clustering multi-view data \cite{MultiNMF,CNMF}. NMF-based multi-view clustering methods have shown to generate superior clustering results which are easy to interpret \cite{MMNMF,UDNMF}. Considering the relations between view-specific (uncorrelated) and common information (correlated), partially shared NMF-based multi-view learning approaches are proposed \cite{PSLF,CoUFC}.
\begin{figure}[t]
\centerline{\includegraphics[width=3in]{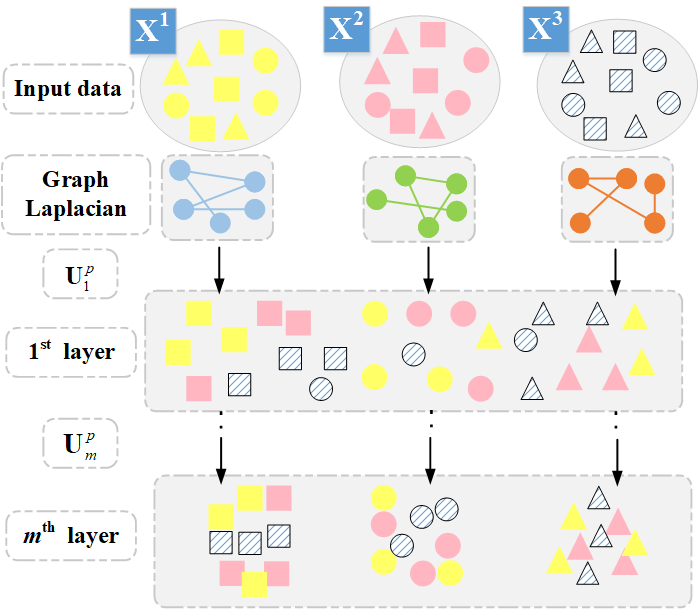}}
\caption{The general framework of the Deep Multi-View Clustering. Data of the same class (\emph{shape}) will become more compact and generate more discriminative representation as the number of decomposition layers increases.}
\label{DMV}
\end{figure}

Nevertheless, most existing methods are single layer structures, which are hard to extract hierarchical structural information \cite{GDNMF} of multi-view data. With the development of deep learning, Wang et al. proposed DCCA to extract the hidden hierarchical information in 2-view data \cite{CCA}. Zhao et al. \cite{DMVC} extended single-view deep matrix factorization \cite{DeepSemiNMF} into multi-view and proposed graph regularized deep multi-view clustering, which can eliminate the interference in multi-view data. Huang et al. \cite{DMD} designed a novel robust deep multi-view clustering model to learn the hierarchical semantics without hyperparameters. For clarity, the general deep matrix factorization framework for multi-view data as illustrated in Fig.~\ref{DMV}. However, most existing deep matrix factorization multi-view methods are merely considering the common information and ignoring the effect of view-specific information in each individual view. Besides, they are formulated as an unsupervised learning problem and inapplicable when partially labeled data is available. Actually, researchers have found that integrating such label information can produce a considerable improvement in learning performance \cite{MvSL,DICS,GPSNMF}.

To this end, we propose a novel deep multi-view clustering methods called Partially Shared Semi-supervised Deep Matrix Factorization (PSDMF). In our method, correlated and uncorrelated features of multi-view data are both considered by partially shared approaches, in which the latent representation of each view is divided into the common part and the view-specific part. A robust sparse regression term with $L_{2,1}$ norm is adopted to integrate the partial label information of the labeled data. Besides, to respect the intrinsic geometric relationship and avoid the parameters problem in different views, we apply graph regularization and auto-weighted strategy to PSDMF. An efficient iterative updating algorithm with pre-training scheme is designed for PSDMF.
We summarize our major contributions as follows:
\begin{itemize}
\item This paper proposes an semi-supervised deep multi-view clustering model to improve the performance of traditional unsupervised deep matrix factorization methods by using label regression learning approach.
\item  To respect the common information and view-specific features of multi-view data, we propose a partially shared deep matrix factorization method to jointly exploit two kinds of information and learn a comprehensive final layer representation of multi-view data.
\item Local invariant graph regularization and the auto-weighted strategy are introduced to preserve the intrinsic geometric structure in each view and further boost the quality of output representation.
\end{itemize}

\begin{figure*}[htbp]
\centerline{\includegraphics[width=6in]{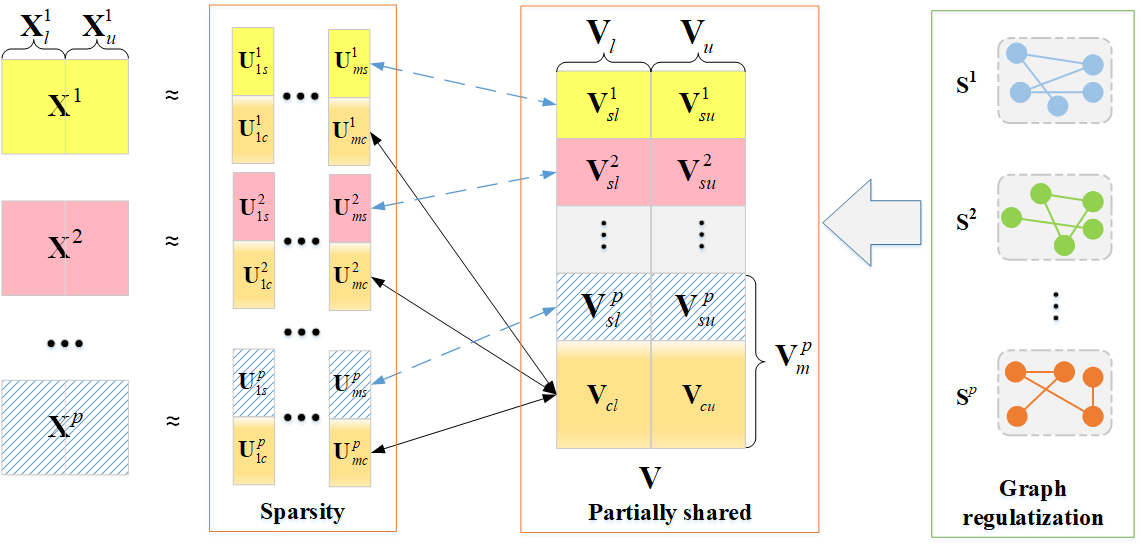}}
\caption{Illustration of the work flow of the proposed Partially Shared Semi-supervised Deep Matrix Factorization.}
\label{flow}
\end{figure*}
The rest of this paper is organized as follows. In Section~\ref{method}, we give a brief review of related works and describe the proposed PSDMF. Section~\ref{opti} present an efficient algorithm to solve the optimization problem. In Section~\ref{exp}, we report the experimental results on five real-world datasets. Finally, Section~\ref{con} concludes this paper. Table. \ref{nota} summarizes the general notations in this article for reader's convenience.
\begin{table}[t]
\label{nota}
\centering
\caption{Notations.}
\begin{tabular}{ll}
\toprule  
Notation& Description \\
\midrule  
$P$& The number of views   \\
$m$& The number of layers \\
$k$& The layer sizes \\
$N$& The number of samples\\
$N_{l}$& The number of labeled samples\\
$N_{u}$& The number of unlabeled samples\\
$K_{s}$ & The dimension of view-specific encoding matrix\\
$K_{c}$ & The dimension of shared encoding matrix\\
$\mathbf{X}^{p}$& The data matrix of the $p$-th view  \\
$\mathbf{U}_{i}^{p}$& The hidden matrix in the $i$-th layer of $p$-th view \\
$\mathbf{V}$& The partially shared factor latent representation matrix \\
$\mathbf{V}_{m}^{p}$& The $m$-th layer partially shared factor of the $p$-th view \\
$\mathbf{W}$ & The regression weight matrix \\
\bottomrule 
\end{tabular}
\label{datasets}
\end{table}

\section{Methodology}\label{method}

\subsection{Overview of Deep Matrix Factorization}
Semi-NMF is not only a useful data dimensionality reduction technique but also beneficial to data clustering \cite{SemiNMF}. Motivated by deep neural network structures, Trigeorgis et al. \cite{DeepSemiNMF} proposed a multi-layer structure Semi-NMF called Deep Semi-NMF to exploit data complex hierarchical information with implicit lower-level hidden attributes. The Deep Semi-NMF decomposes dataset $\mathbf{X}$ hierarchically, and the process can be formulated as:
\begin{equation}\begin{aligned}
\mathbf{X} & \approx \mathbf{U}_{1} \mathbf{V}_{1}^{+} \\
\mathbf{X} & \approx \mathbf{U}_{1} \mathbf{U}_{2} \mathbf{V}_{2}^{+} \\
  & \vdots \\
\mathbf{X} & \approx \mathbf{U}_{1} \ldots \mathbf{U}_{m} \mathbf{V}_{m}^{+}
\end{aligned}\end{equation}
where $m$ is the numbers of layers, $\mathbf{U}_{i}$ is the $i$-th layer hidden matrix, $\mathbf{V}_{i}^{+}$ denotes the $i$-th layer low-dimensional representation matrix and we use the notation $+$ to state that a matrix contains only non-negative elements. The loss function of Deep Semi-NMF is:

\begin{equation}
\min _{\mathbf{U}_{i}, \mathbf{V}_{m}} \left\|\mathbf{X}-\mathbf{U}_{1}\cdot\cdot\cdot \mathbf{U}_{m} \mathbf{V}_{m}\right\|_{F}^{2} \text { s.t. } \mathbf{V}_{m} \geq 0.
\end{equation}

The Semi-NMF and Deep Semi-NMF mentioned above can be regarded as single view algorithms. For multi-view data, Zhao et al. \cite{DMVC} presented a new Deep Semi-NMF framework with graph regularization (DMVC) which can eliminate the negative interference in the multi-source data and obtain an effective consensus representation in the final layer. Given data $\mathbf{X}$ consisting of $P$ views, denoted by $\mathbf{X}=\{\mathbf{X}^{1}, \mathbf{X}^{2},...,\mathbf{X}^{P}\}$, the loss function of DMVC is:

\begin{equation}
\begin{aligned}
\min _{\mathbf{Z}_{i}^{p}, \mathbf{H}_{i}^{p},\mathbf{H}_{m}, \alpha^{p}}& \sum_{p=1}^{P}\left(\alpha^{p}\right)^{r}\left(\left\|\mathbf{X}^{p}-\mathbf{U}_{1}^{p} \mathbf{U}_{2}^{p} \ldots \mathbf{U}_{m}^{p} \mathbf{V}_{m}\right\|_{\mathrm{F}}^{2}\right. \\
&\left.\quad +\beta \operatorname{tr}\left(\mathbf{V}_{m} \mathbf{L}^{p} \mathbf{V}_{m}^{T}\right)\right) \\
\text { s.t. }& \mathbf{V}_{i}^{p} \geq 0, \mathbf{V}_{m} \geq 0, \sum_{p=1}^{P} \alpha^{p}=1, \alpha^{p} \geq 0
\end{aligned}
\end{equation}
where $\alpha^{p}$ denotes the weighting coefficient of the $p$-th view and $r$ is a important hyperparameter to control the weights distribution. $\mathbf{L}^{p}$ is the graph Laplacian of the graph for view $p$ and $\beta$ is used to adjust the contribution of the graph constrains. The details of how to construct a graph matrix will be discussed in the next subsection. However, as an unsupervised method, DMVC cannot make use of partial prior knowledge of data (e.g. labels). Besides, DMVC only concerns common representation of multi-view data and ignores the view-specific features. In this paper, we propose a novel semi-supervised deep matrix factorization model to address such challenging problems.

\subsection{Partially Shared Semi-supervised Deep Matrix Factorization}

To make full use of priori knowledge and learn low-dimensional factors with powerful discrimination, motivated by recently proposed label regression learning technique \cite{PSLF,GPSNMF}, we combine the following $L_{2,1}$-norm regularized regression in our model:
 \begin{equation}
\label{regre}
\min \left\|\mathbf{W}^{T} \mathbf{V}_{l}-\mathbf{Y}\right\|_{F}^{2}+\gamma\|\mathbf{W}\|_{2,1}.
\end{equation}
where $\mathbf{V}_{l} \in \mathbb{R}^{K\times N_{l}}$ denotes the labeled part of representation matrix $\mathbf{V}$ and $\mathbf{V}_{u}\in \mathbb{R}^{K\times N_{u}}$ is the unlabeled part as shown in Fig.~\ref{flow} (i.e., $\mathbf{V}=[\mathbf{V}_{l},\mathbf{V}_{u}]$). $\mathbf{W}\in \mathbb{R}^{K\times C}$ is the regression coefficient matrix and the $L_{2,1}$-norm is used to enforce it sparse in rows.
Consider we have a priori knowledge that those samples share the same label, a binary weight matrix $\mathbf{Y}\in \mathbb{R}^{C\times N}$ is constructed by following rule:
\begin{equation}
\label{Y}
\mathbf{Y}_{cn}=
\left\{\begin{array}
{ll}{1} & {\text { if } n\text {-th data point belongs to the }c\text {-th class}} \\
    {0} & {\text { otherwise }}
\end{array}\right.
\end{equation}

The effectiveness of graph regularization technique has been shown in recent research work \cite{GNMF,DMVC} and it is able to keep the geometric structure of data when the dimension is changed. Similar to the DMVC, we construct a local graph Laplacian matrix to preserve the local geometrical structure of the each view $\mathbf{X}^{p}$. As introduced in \cite{DMVC}, the graph binary weight matrix $\mathbf{S}^{p}$ is constructed in \emph{k}-nearest neighbor (\emph{k}-NN) fashion. Formally, the regularization term $\mathcal{R}^{p}$ is calculated:
\begin{equation}
\label{L}
\begin{aligned} \mathcal{R}^{p} &=\frac{1}{2} \sum_{j, q=1}^{n}\left\|\mathbf{v}_{j}^{p}-\mathbf{v}_{q}^{p}\right\|^{2} \mathbf{S}_{j q}^{p} \\ &=\sum_{j=1}^{n} \mathbf{v}_{j}^{p} \mathbf{v}_{j}^{p} \mathbf{D}_{j j}^{p}-\sum_{j, q=1}^{n} \mathbf{v}_{j}^{p} \mathbf{v}_{l}^{p} \mathbf{S}_{j q}^{p} \\ &=\operatorname{Tr}\left(\mathbf{V}^{p} \mathbf{D}^{p} (\mathbf{V}^{p})^{T}\right)-\operatorname{Tr}\left(\mathbf{V}^{p} \mathbf{S}^{p} (\mathbf{V}^{p})^{T}\right)\\
&=\operatorname{Tr}\left(\mathbf{V}^{p} \mathbf{L}^{p} (\mathbf{V}^{p})^{T}\right) \end{aligned}
\end{equation}
where $\mathbf{D}_{jj}^{p}=\sum_{q}\mathbf{S}_{jq}^{p}$. $\mathbf{L}^{p} = \mathbf{D}^{p}-\mathbf{S}^{p}$ denotes the graph Laplacian matrix for each view data.

Different from the existing deep matrix factorization multi-view methods, we use partially shared strategy to jointly exploit view-specific and common features. The partially shared latent representation matrix $\mathbf{V}\in \mathbb{R}^{K\times N}$ and $K= K_{s}\times P+K_{c}$. The common factor ratio $\lambda = K_{c}/(K_{s}+K_{c})$. The final $m$-th layer partially shared factor $\mathbf{V}_{m}^{p}\in \mathbb{R}^{(K_{s}+K_{c})\times N}$ is divided into four parts: the labeled and unlabeled view-specific encoding matrix (i.e. $\mathbf{V}_{sl}^{p}\in \mathbb{R}^{K_{s}\times N_{l}}$ and $\mathbf{V}_{su}^{p}\in \mathbb{R}^{K_{s}\times N_{u}}$), the labeled and unlabeled shared encoding matrix (i.e. $\mathbf{V}_{cl}\in \mathbb{R}^{K_{c}\times N_{l}}$ and $\mathbf{V}_{cu}\in \mathbb{R}^{K_{c}\times N_{u}}$). We propose a general partially shared deep matrix factorization framework as follows:
\begin{equation}\label{partshare}\begin{aligned}
\min \sum_{p=1}^{P}\left\|\mathbf{X}^{p}-
\left[
\mathbf{U}_{1s}^{p},
\mathbf{U}_{1c}^{p}
\right]
\cdot\cdot\cdot
\left[
\mathbf{U}_{ms}^{p},
\mathbf{U}_{mc}^{p}
\right]
\left[\begin{array}{c}
\mathbf{V}_{sl}^{p}, \mathbf{V}_{su}^{p} \\
\mathbf{V}_{cl}, \mathbf{V}_{cu}
\end{array}\right]\right\|_{F}^{2} \\
\text { s.t. } \mathbf{V}_{sl}^{p}, \mathbf{V}_{su}^{p},\mathbf{V}_{cl}, \mathbf{V}_{cu} \geq 0
\end{aligned}\end{equation}
where $\mathbf{U}_{is}^{p}$ and $\mathbf{U}_{ic}^{p}$ denote the hidden view-specific matrix and common matrix, respectively. To simplify the problem, we use $\mathbf{V}_{m}^{p} = [\mathbf{V}_{s}^{p};\mathbf{V}_{c}] = \left[[\mathbf{V}_{sl}^{p}, \mathbf{V}_{su}^{p}]; [\mathbf{V}_{cl},\mathbf{V}_{cu}]\right]$, $\mathbf{U}_{i}^{p} = [\mathbf{U}_{is}^{p};\mathbf{U}_{ic}^{p}]$, by combining Eq.(\ref{regre}), Eq.(\ref{L}) and Eq.(\ref{partshare}), we get the cost function of PSDMF as:
\begin{equation}
\label{PSDMF}
\begin{aligned}
\min _{\mathbf{U}_{i}^{p}, \mathbf{V}_{m}^{p}, \mathbf{W}} O=& \sum_{p=1}^{P}\left(\alpha^{p}\left\|\mathbf{X}^{p}-\mathbf{U}_{1}^{p}\cdot\cdot\cdot \mathbf{U}_{m}^{p} \mathbf{V}_{m}^{p}\right\|_{F}^{2}\right.\\
&\left.+\mu \operatorname{tr}\left(\mathbf{V}_{m}^{p} \mathbf{L}^{p}\left(\mathbf{V}_{m}^{p}\right)^{T}\right)\right) \\
&+\beta\left(\left\|\mathbf{W}^{T} \mathbf{V}_{l}-\mathbf{Y}\right\|_{F}^{2}+\gamma\|\mathbf{W}\|_{2,1}\right) \\
& \text { s.t. } \mathbf{V}_{m}^{p} \geq 0,\forall p
\end{aligned}\end{equation}
where $\alpha^{p}$ is the weighting coefficient for the $p$-th view and it has a great influence on the effect of the model. In DMVC \cite{DMVC}, the weights distribution smooth is determined by parameter $r$, but $r$ needs to be searched in a large range manually, which makes it to adjust. To avoid this problem, inspired by \cite{AWMVL,SWMKL}, we use the auto-weighted strategy that obtains the value of $\alpha^{p}$ based on the distance between the data and the decomposition matrices, as follows:
\begin{equation}
\label{alph}
\alpha^{p} = \frac{1}{2\left\|\mathbf{X}^{p}-\mathbf{U}_{1}^{p}\cdot\cdot\cdot \mathbf{U}_{m}^{p} \mathbf{V}_{m}^{p}\right\|_{F}}
\end{equation}
\section{Optimization}\label{opti}
To expedite the approximation of the matrices in our proposed model, we conduct the pre-training \cite{MCL} by decomposing each view $\mathbf{X}^{p}\approx \mathbf{U}_{1}^{p} \mathbf{V}_{1}^{p}$ by minimizing the Semi-NMF $\left\| \mathbf{X}^{p}-\mathbf{U}_{1}^{p}\mathbf{V}_{1}^{p}\right\|_{F}^{2}$,
where $\mathbf{U}_{1}^{p} \in \mathbb{R}^{M^{p}\times k_{1}}$ and $\mathbf{V}_{1}^{p} \in \mathbb{R}^{k_{1}\times N}$. Then the factor matrix $\mathbf{V}_{1}^{p}$ is further decomposed as $\mathbf{V}_{1}^{p}\approx \mathbf{U}_{2}^{p} \mathbf{V}_{2}^{p}$ by minimizing the $\left\| \mathbf{V}_{2}^{p}-\mathbf{U}_{2}^{p}\mathbf{V}_{2}^{p}\right\|_{F}^{2}$, where $\mathbf{U}_{2}^{p} \in \mathbb{R}^{k_{1} \times k_{2}}$, $\mathbf{V}_{1}^{p} \in \mathbb{R}^{k_{2}\times N}$ and $k_{1},k_{2}$ denote the dimensions for layer 1 and layer 2. The process will be repeated till all of the layers are pre-trained. The optimization of Semi-NMF can be derived following a similar process as described in~\cite{SemiNMF}. To save the space, we omit the updating rules here.

\textbf{Update rule for hidden matrix $\mathbf{U}_{i}^{p}$:}
By fixing $\mathbf{V}_{m}^{p}$ and $\mathbf{W}$, we minimize the objective function~(\ref{PSDMF})  with respect to $\mathbf{U}_{i}^{p}$. Letting the partial derivative $\partial O / \mathbf{U}_{i}^{p}=0$,
we can obtain
\begin{equation}
\label{Ui}
\mathbf{U}_{i}^{p}=\left(\Phi_{i-1}^{T} \Phi_{i-1}\right)^{-1}\left( \Phi_{i-1}^{T} \mathbf{X}^{p} \mathbf{U}_{im}^{T}\right) \\
\left(\mathbf{U}_{im} \mathbf{U}_{im}^{T}\right)^{-1}
\end{equation}
where $\Phi_{i-1} = \mathbf{U}_{1}^{p} \mathbf{U}_{2}^{p} \ldots \mathbf{U}_{i-1}^{p}$ and $\mathbf{U}_{im} =
\mathbf{U}_{i+1}^{p} \ldots \mathbf{U}_{m}^{p} \mathbf{V}_{m}^{p}$.

\textbf{Update rule for regression weight matrix $\mathbf{W}$:}
Following~\cite{L21}, the derivative of objective function~(\ref{PSDMF}) with respect to $\mathbf{W}$ is as follows:
\begin{equation}
\frac{\partial O}{\mathbf{W}}=2\beta\left( \mathbf{V}_{l}(\mathbf{V}_{l}^{T}\mathbf{W}-\mathbf{Y}^{T})+\gamma \mathbf{E}\mathbf{W}\right)
\end{equation}
 where $\mathbf{E}$ is a diagonal matrix with $e_{ii}=1/2\|\mathbf{w}_{i}\|_{2}$. According to the optimization theory, set $\partial O/\mathbf{W} = 0$, then we can obtain the update rule for $\mathbf{W}$ is
 \begin{equation}
 \label{W}
\mathbf{W}=\left( \mathbf{V}_{l}\mathbf{V}_{l}^{T}+\gamma \mathbf{E}\right)^{-1}\mathbf{V}_{l}\mathbf{Y}^{T}.
\end{equation}

\textbf{Update rule for partially shared factor matrix $\mathbf{V}_{m}^{p}$:} As illustrated in Fig.~\ref{flow}, $\mathbf{V}_{m}^{p}$ can be divided into four parts: $\mathbf{V}_{sl}^{p},\mathbf{V}_{su}^{p},\mathbf{V}_{cl}$ and $\mathbf{V}_{cu}$. Similarly, each data matrix $\mathbf{X}^{p}$ is divided into two parts $\mathbf{X}_{l}^{p}$ and $\mathbf{X}_{u}^{p}$; each Laplacian Graph matrix $\mathbf{L}^p$ is divided into two parts $\mathbf{L}_{l}^p$ and $\mathbf{L}_{u}^p$. \par
For the constraint $\mathbf{V}_{m}^{p} \geq 0$, we introduce the Lagrangian multiplier $\eta$ as follows:
\begin{equation}\begin{aligned}
\mathcal{L}\left(\mathbf{V}_{m}^{p}\right) &=\sum_{p=1}^{P}\left(\alpha^{p}\left\|\mathbf{X}^{p}- \Phi_{m} \mathbf{V}_{m}^{p}\right\|_{\mathrm{F}}^{2}+\mu \operatorname{tr}\left(\mathbf{V}_{m}^{p} \mathbf{L}^{p} (\mathbf{V}_{m}^{p})^{T}\right)\right.\\
&\left.+\beta \left\|\mathbf{W}^{T} \mathbf{V}_{l}-\mathbf{Y} \right\|-tr(\eta^{T} \mathbf{V}_{m}^{p})\right)
\end{aligned}\end{equation}
where $\Phi_{m} = \mathbf{U}_{1}^{p} \mathbf{U}_{2}^{p} \ldots \mathbf{U}_{m}^{p}$. Accordingly, $\Phi_{m}$ is divided into two parts $\Phi_{ms}$ and $\Phi_{mc}$. For the convenience of writing, we denote that $\mathbf{A}^{p} = [\Phi_{ms}, \Phi_{mc}] [\mathbf{V}_{sl}^{p};\mathbf{V}_{cl}] $, $\mathbf{B}^{p} = [\Phi_{ms}, \Phi_{mc}] [\mathbf{V}_{su}^{p}; \mathbf{V}_{cu}] $. \par
The zero gradient condition of $\mathcal{L}\left(\mathbf{V}_{m}^{p}\right)$ with respect to $\mathbf{V}_{sl}^{p},\mathbf{V}_{su}^{p},\mathbf{V}_{cl}$ and $\mathbf{V}_{cu}$, respectively, we have
\begin{equation}\begin{aligned}
\frac{\partial \mathcal{L}}{\partial \mathbf{V}_{sl}^{p}}=&  \alpha^{p}\Phi_{ms}^{T}\left(\mathbf{A}^{p}-\mathbf{X}_{l}^{p}\right)+\mu \mathbf{V}_{s}^{p}\mathbf{L}_{l}^{p}+ \beta \mathbf{F}_{s}^{p}-\eta_{sl}=0 \\
\frac{\partial \mathcal{L}}{\partial \mathbf{V}_{su}^{p}}=&  \alpha^{p}\Phi_{ms}^{T}\left(\mathbf{B}^{p}-\mathbf{X}_{u}^{p}\right)+\mu\mathbf{V}_{s}^{p}\mathbf{L}_{u}^{p}-\eta_{su}=0 \\
\frac{\partial \mathcal{L}}{\partial \mathbf{V}_{c1}}=& \sum_{p=1}^{P}  \alpha^{p}\Phi_{mc}^{T}\left(\mathbf{A}^{p}-\mathbf{X}_{l}^{p}\right)+\mu \mathbf{V}_{c}^{p}\mathbf{L}_{l}^{p}
+ \beta \mathbf{F}_{c}-\eta_{cl}=0 \\
\frac{\partial \mathcal{L}}{\partial \mathbf{V}_{cu}}=& \sum_{p=1}^{P}  \alpha^{p}\Phi_{mc}^{T}\left(\mathbf{B}^{p}-\mathbf{X}_{u}^{p}\right)+\mu\mathbf{V}_{c}^{p}\mathbf{L}_{u}^{p}
-\eta_{cu}=0
\end{aligned}\end{equation}
where $\mathbf{F}=\mathbf{W}(\mathbf{W}^{T} \mathbf{V}_{l}- \mathbf{Y})= [\mathbf{F}_{s}^{1};...; \mathbf{F}_{s}^{p}; \mathbf{F}_{c}]$. Following a similar proof to~\cite{DMVC}, using the Karush-Kuhn-Tucker condition, we can formulate the updating rules for $\mathbf{V}_{m}^{p}$:
\begin{equation}
\label{Vm1}
\begin{aligned}
&\mathbf{V}_{sl}^{p} = \mathbf{V}_{sl}^{p}\odot \\ &\sqrt{\frac{\alpha^{p}[\Phi_{ms}^{T}\mathbf{A}^{p}]^{-}+\alpha^{p}[\Phi_{ms}^{T}\mathbf{X}_{l}^{p}]^{+}+\mu [\mathbf{V}_{s}^{p}\mathbf{L}_{l}^{p}]^{-}+ \beta [\mathbf{F}_{s}^{p}]^{-}}{\alpha^{p}[\Phi_{ms}^{T}\mathbf{A}^{p}]^{+}+\alpha^{p}[\Phi_{ms}^{T}\mathbf{X}_{l}^{p}]^{-}+\mu [\mathbf{V}_{s}^{p}\mathbf{L}_{l}^{p}]^{+}+ \beta [\mathbf{F}_{s}^{p}]^{+}}}\\
&\mathbf{V}_{su}^{p} = \mathbf{V}_{su}^{p}\odot \\ &\sqrt{\frac{\alpha^{p}[\Phi_{ms}^{T}\mathbf{B}^{p}]^{-}+\alpha^{p}[\Phi_{ms}^{T}\mathbf{X}_{u}^{p}]^{+}+\mu [\mathbf{V}_{s}^{p}\mathbf{L}_{u}^{p}]^{-}}{\alpha^{p}[\Phi_{ms}^{T}\mathbf{B}^{p}]^{+}+\alpha^{p}[\Phi_{ms}^{T}\mathbf{X}_{u}^{p}]^{-}+\mu [\mathbf{V}_{s}^{p}\mathbf{L}_{u}^{p}]^{+}}}\\
&\mathbf{V}_{cl} = \mathbf{V}_{cl}\odot \\ &\sqrt{\frac{\sum\limits_{p=1}^{P}\alpha^{p}\left([\Phi_{mc}^{T}\mathbf{A}^{p}]^{-}+[\Phi_{ms}^{T}\mathbf{X}_{l}^{p}]^{+}\right)+\mu [\mathbf{V}_{c}^{p}\mathbf{L}_{l}^{p}]^{-}+ \beta [\mathbf{F}_{c}]^{-}}{\sum\limits_{p=1}^{P}\alpha^{p}\left([\Phi_{ms}^{T}\mathbf{A}^{p}]^{+}+[\Phi_{ms}^{T}\mathbf{X}_{l}^{p}]^{-}\right)+\mu [\mathbf{V}_{c}^{p}\mathbf{L}_{l}^{p}]^{+}+ \beta [\mathbf{F}_{c}]^{+}}}\\
&\mathbf{V}_{cu} = \mathbf{V}_{cu}\odot \\ &\sqrt{\frac{\sum\limits_{p=1}^{P}\alpha^{p}[\Phi_{mc}^{T}\mathbf{B}^{p}]^{-}+\sum\limits_{p=1}^{P}\alpha^{p}[\Phi_{ms}^{T}\mathbf{X}_{u}^{p}]^{+}+\mu [\mathbf{V}_{c}^{p}\mathbf{L}_{u}^{p}]^{-}}{\sum\limits_{p=1}^{P}\alpha^{p}[\Phi_{ms}^{T}\mathbf{B}^{p}]^{+}
+\sum\limits_{p=1}^{P}\alpha^{p}[\Phi_{ms}^{T}\mathbf{X}_{u}^{p}]^{-}+\mu [\mathbf{V}_{c}^{p}\mathbf{L}_{u}^{p}]^{+}}},\\
\end{aligned}\end{equation}
where $[\mathbf{H}]^{+}$ and $[\mathbf{H}]^{-}$ denote a matrix that all the negative elements are replaced by $0$ and all the positive elements are replaced by $0$, respectively. That is,
\begin{equation}
\forall k,j [\mathbf{H}]_{kj}^{+}=\frac{|\mathbf{H}_{kj}|+\mathbf{H}_{kj}}{2}, [\mathbf{H}]_{kj}^{-}=\frac{|\mathbf{H}_{kj}|-\mathbf{H}_{kj}}{2}.
\end{equation}

\begin{algorithm} 
\caption{Optimization algorithm of PSDMF} 
\label{alg} 
\begin{algorithmic}[1] 
\REQUIRE Multi-view Data $\{\mathbf{X}^{p}\}_{p=1}^{P}$, parameters $\mu ,\beta,\gamma,\lambda,K$,  layer sizes $k$
\ENSURE Partially shared latent representation $\mathbf{V}$, the regression weight matrix $\mathbf{W}$ 
\STATE \textbf{Initialize}
\STATE Construct partially label matrix $\mathbf{Y}$ via Eq. (\ref{Y})
\FORALL{layers in each view}
\STATE $(\mathbf{U}_{i}^{p}, \mathbf{V}_{i}^{p})$$\leftarrow$ Semi-NMF$(\mathbf{V}_{i-1}^{p}, \boldsymbol{k}_{i})$
\ENDFOR
\WHILE{ not converged}
\STATE Update $\mathbf{W}$ via Eq. (\ref{W})
\FOR{$p =1$ to $P$}
\STATE Update $\alpha^{p}$ via Eq. (\ref{alph})
\FOR{$i =1$ to $m$}
\STATE Update $\mathbf{U}_{i}^{p}$ via Eq. (\ref{Ui})
\ENDFOR
\STATE Update $\mathbf{V}_{m}^{p}$ via Eq. (\ref{Vm1})
\ENDFOR
\ENDWHILE
\end{algorithmic}
\end{algorithm}

\begin{table*}[htbp]
	\centering
	\caption{Results on five datasets( mean $\pm$ standard deviation). Higher value indicates better performance and the highest values are in boldface.}
    \label{YaleB}
	\begin{tabular}{p{1.1cm}p{0.5cm}ccccccccp{1cm}}
		\toprule  
	 Datasets &  & DMVC&GMC &lLSMC&LMVSC&DICS&MvSL&PSLF&GPSNMF&Ours\\
		\midrule  
Extended  &$\textrm{ACC}$&50.34$\pm$0.07&43.38$\pm$0.00&53.41$\pm$1.29		
&36.62$\pm$0.00&47.54$\pm$3.58&19.64$\pm$1.81&43.79$\pm$4.90&69.56$\pm$8.04&\textbf{87.38$\pm$3.45}		\\
   Yale B  &$\textrm{NMI}$& 49.97$\pm$0.14&44.90$\pm$0.00&53.40$\pm$0.64&28.09$\pm$0.00 &50.05$\pm$5.11&11.82$\pm$2.88&32.14$\pm$4.85&60.44$\pm$9.12&\textbf{83.53$\pm$2.44}  \\
            &$\textrm{Purity}$&50.49$\pm$0.07&43.69$\pm$0.00&53.45$\pm$1.26&42.77$\pm$0.00 &51.39$\pm$3.82&20.52$\pm$1.66&44.10$\pm$4.36&69.56$\pm$8.04&\textbf{87.38$\pm$3.45} \\
            \midrule  
 Prokaryotic  &$\textrm{ACC}$&53.50$\pm$2.23&49.55$\pm$0.00&52.31$\pm$6.01&57.53$\pm$0.00		&56.07$\pm$11.76&35.68$\pm$2.51		&46.49$\pm$6.63& 63.45$\pm$2.59&\textbf{66.83$\pm$2.94}\\
             &$\textrm{NMI}$&3.28$\pm$0.66&	19.34$\pm$0.00&16.49$\pm$8.59&13.37$\pm$0.00&12.63$\pm$9.88	&1.05$\pm$0.41&1.69$\pm$1.02&18.46$\pm$5.48&\textbf{21.08$\pm$4.21}  \\
            &$\textrm{Purity}$&57.31$\pm$0.65&	58.44$\pm$0.00&63.08$\pm$4.15&62.94$\pm$0.00 &60.14$\pm$5.16&57.03$\pm$0.07&57.46$\pm$0.73&63.45$\pm$2.59&\textbf{66.53$\pm$2.94} \\
              \midrule  
 Caltech101  &$\textrm{ACC}$&54.60$\pm$0.43&69.20$\pm$0.00&51.98$\pm$2.61&69.47$\pm$0.00&50.95$\pm$10.61&46.42$\pm$4.14	&76.88$\pm$3.09 &  89.77$\pm$2.54&\textbf{90.24$\pm$3.08}				\\
  -7 &$\textrm{NMI}$&37.60$\pm$5.79&65.95$\pm$0.00&53.22$\pm$2.22&46.68$\pm$0.00 &53.08$\pm$7.71&51.59$\pm$2.81&40.94$\pm$7.75&74.53$\pm$4.10&\textbf{75.52$\pm$5.36}  \\
            &$\textrm{Purity}$&79.55$\pm$4.64&88.47$\pm$0.00&85.61$\pm$1.64&77.07$\pm$0.00 &85.28$\pm$3.70&83.73$\pm$1.49&77.56$\pm$4.12&90.38$\pm$2.33&\textbf{91.12$\pm$2.03} \\
              \midrule  
 Caltech101  &$\textrm{ACC}$&51.72$\pm$1.72&45.64$\pm$0.00&45.03$\pm$4.16&42.08$\pm$0.00&35.75$\pm$3.83	&42.01$\pm$2.86		&55.75$\pm$10.35&  77.33$\pm$3.34&\textbf{79.74$\pm$3.73}			\\
  -20 &$\textrm{NMI}$&52.18$\pm$1.49&48.09$\pm$0.00&61.21$\pm$1.14&49.64$\pm$0.00 &55.99$\pm$2.58&60.48$\pm$1.45	&44.85$\pm$5.77&67.48$\pm$3.52&\textbf{71.24$\pm$3.06}  \\
            &$\textrm{Purity}$&70.23$\pm$0.92&55.49$\pm$0.00&76.95$\pm$0.79& 50.84$\pm$0.00&69.32$\pm$2.47&76.16$\pm$1.17&62.55$\pm$4.94&77.81$\pm$2.60&\textbf{79.74$\pm$2.84} \\
               \midrule  
 MSRCV1  &$\textrm{ACC}$&40.76$\pm$2.37&74.76$\pm$0.00&69.91$\pm$1.79&64.76$\pm$0.00		&51.91$\pm$6.61&78.07$\pm$5.83		&82.96$\pm$4.51& 72.17$\pm$4.03&\textbf{86.24$\pm$5.69}				 \\
   &$\textrm{NMI}$&27.60$\pm$3.17&	70.09$\pm$0.00&62.37$\pm$2.71&58.81$\pm$0.00&50.48$\pm$11.29&68.32$\pm$4.69&75.04$\pm$3.49&67.41$\pm$4.34&\textbf{76.14$\pm$3.42}   \\
            &$\textrm{Purity}$&45.14$\pm$2.41&	79.05$\pm$0.00&72.86$\pm$1.38&69.05$\pm$0.00&58.09$\pm$8.68&78.52$\pm$5.73&82.96$\pm$4.51&74.92$\pm$4.44&\textbf{86.24$\pm$4.45} \\
		\bottomrule  
	\end{tabular}
\end{table*}

Until now, we have all the update rules done. Updates iterations repeatedly until convergence. The overall optimization process of PSDMF is outlined in Algorithm~\ref{alg}, where the "Semi-NMF" procedure performs the pre-train as described earlier. Once the optimal partially shared latent representation matrix $\mathbf{V}$ and regression weight matrix $\mathbf{W}$  are obtained, cluster label $y = \arg\max_{c}y_{c,i}$ where $\boldsymbol{y}_{i} = \mathbf{W}^{T}\boldsymbol{v}_{i}$.
\section{Experiments}\label{exp}

To comparatively study the performance of PSDMF, we consider several state-of-the-art baselines as stated bellow:\\
Multi-view clustering via deep matrix factorization (\textbf{DMVC}) \cite{DMVC}, graph-based multi-view clustering (\textbf{GMC}) \cite{GMC}, latent multi-view subspace clustering (\textbf{lLSMC}) \cite{lLSMC}, large-scale multi-view subspace clustering (\textbf{LMVSC}) \cite{LMVSC}, multi-view discriminative learning via joint non-negative matrix factorization (\textbf{DICS}) \cite{DICS}, graph-regularized multi-view semantic subspace learning (\textbf{MvSL}) \cite{MvSL}, partially shared latent factor learning (\textbf{PSLF}) \cite{PSLF} and semi-supervised multi-view clustering with graph-regularized partially shared non-negative matrix factorization (\textbf{GPSNMF}) \cite{GPSNMF}. Among them, four methods are unsupervised (i.e., DMVC, GMC, lLSMC, LMVSC) and the others are semi-supervised.
\subsection{Datasets and Evaluation Metric }
We perform experiments on five benchmark datasets: Extended Yale B, Prokaryotic, MSRCV1, Caltech101-7 and 20. More detailed information about these data is shown in Table~\ref{datasets}.
 \begin{table}[t]
\centering
\caption{Statistics of datasets used in experiments.}
\begin{tabular}{ccp{1cm}p{1cm}}
\toprule  
Dataset& Dimension of Views& Sizes & Classes\\
\midrule  
Extended Yale B& 2500/3304/6750& 650&10 \\
Prokaryotic& 393/3/438& 551&4 \\
Caltech101-7/20& 48/40/254/1984/512/928&  1474/2386&7/20 \\
MSRCV1 &24/576/512/256/254 & 210&7\\
\bottomrule 
\end{tabular}
\label{datasets}
\end{table}

For evaluation metrics, we use Accuracy (\textbf{ACC}), Normalized Mutual Information (\textbf{NMI}) and \textbf{Purity} to comprehensively measure the clustering performance in our experiments. The formal definitions of these metrics, omitted here to save space, can be found in \cite{UDNMF}.
\subsection{Experiment Setup}

Following \cite{GPSNMF}, for PSLF, GPSNMF and PSDMF, the dimension of partially shared latent representation $K$ and the common factor ratio $\lambda$ are set to 100 and 0.5, respectively. Thus, $K_{c} + K_{s}\times P = 100$ and  $K_{c}/(K_{s}+K_{c})=0.5$. In addition, they all predicate the labels with regression matrix $\mathbf{W}$. Other than these three algorithms, all the above methods are evaluated by the classic clustering algorithm, K-means. In terms of semi-supervised methods (i.e. DICS, MvSL, PSLF, GPSNMF and PSDMF), we denote $10\%$ as the proportions of the labeled data. Following \cite{DMVC}, the layer sizes of deep structure models (i.e. DMVC, PSDMF) are set as $[100, 50]$.

For all compared methods, 10 repeated times are run to reduce the randomness caused by initialization, and report their average scores and standard deviations. As to the compared methods, source codes are obtained from the authors' websites, and we set their parameters to the optimal value if they have.
\subsection{Performance Evaluation}
The performances of multi-view clustering is shown in Table~\ref{YaleB}. From the experiment results, we have drawn some conclusions as following: In general, semi-supervised algorithms are superior to unsupervised algorithms. For example, our PSDMF significantly improves the performance compared with DMVC all over the five datasets, even when only $10\%$ label information is used. For the dataset Extended Yale B, we raise the performance bar by around $18\%$ in ACC, $13\%$ in NMI and $18\%$ in Purity. On average, we improve the state-of-the-art GPSNMF by more than $16\%$, which means multi-layer structure methods have clustering advantages over single-layer methods.
Through the partially shared deep representation, PSDMF eliminates the influence of undesirable factors by considering both view-specific and common features, and keeps the important information in the final representation layer. In different datasets experiments, our model achieves better performances than the state-of-the-art methods, which demonstrate the robustness of ours.
\begin{figure}[t]
\centerline{\includegraphics[width=3.5in]{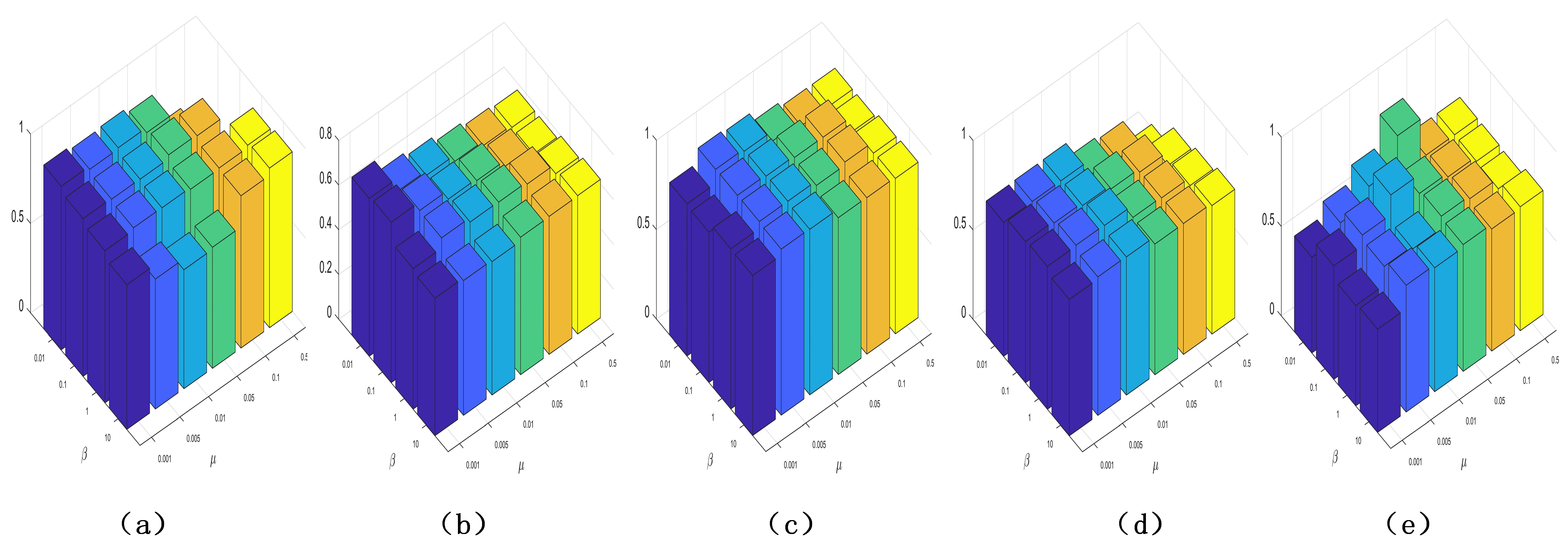}}
\caption{The performance of PSDMF in terms of \textbf{ACC} with different value of parameters $\mu$ and $\beta$. (a) Extended Yale B; (b) Prokaryotic; (c) Caltech101-7; (d) Caltech101-20; (e) MSRCV1.}
\label{para}
\end{figure}

\subsection{Parameters Sensitivity}
There are three nonnegative essential parameters (i.e. $\mu,\beta,\gamma$) in our proposed model. $\mu$ controls the smoothness of the partially shared factor. $\gamma$ and $\beta$ controls the sparsity of the weight matrix $\mathbf{W}$ and balances the relation between the regression term and data reconstruction, respectively. Following \cite{PSLF,GPSNMF}, $\gamma=10$ usually obtains better performance. As is shown in Fig.~\ref{para}, we test the sensitivity of the parameters $\mu$ and $\beta$ in terms of clustering results, which are tuned in the range of $\{0.001, 0.005, 0.01, 0.05, 0.1, 0.5\}$ and $\{0.01, 0.1 ,1, 10\}$, respectively. We can observe that the parameters affect the experimental results in different datasets, which indicates that $\mu$ and $\beta$ play an important role in PSDMF. In practice, we fix $\mu=0.1$ and $\beta=10$ as default in our experiments.

\section{conclusion}\label{con}
In this paper, we introduced a novel semi-supervised deep matrix factorization model for multi-view learning, called PSDMF, which is able to learn a comprehensive partially shared latent final-layer representation for multi-view data. Through the partially shared multi-layer matrix factorization structure, our method is capable of exploiting view-specific and common features among different views, simultaneously. Benefitting from the label regression term, it can incorporate information from the labeled data. Furthermore, we utilize graph regularization and auto-weighted strategy to preserve the intrinsic geometric structure of the data. An iterative optimization algorithm is developed to deal with the PSDMF. Experimental results show that the proposed model achieves superior performances.

\section*{Acknowledgment}
This work was supported by the National Natural Science Foundation of China under Grant Nos. 61722304,
61801133, and 61803096, in part by the Guangdong Science and Technology Foundation under Grant Nos.
2019B010118001, 2019B010121001, and 2019B010154002, National Key Research and Development Project,
China under Grant No. 2018YFB1802400.

\vspace{12pt}

\end{document}